\documentclass[conference]{IEEEtran}
\IEEEoverridecommandlockouts

\usepackage{cite}
\usepackage{amsmath,amssymb,amsfonts}
\usepackage{algorithmic}
\usepackage{graphicx}
\usepackage{textcomp}
\usepackage{xcolor}
\usepackage{multirow}
\usepackage{booktabs}
\usepackage{float}
\usepackage{array}
\usepackage{threeparttable}
\usepackage{makecell}
\usepackage[table]{xcolor}
\usepackage{url}
\usepackage{graphicx}
\def\BibTeX{{\rm B\kern-.05em{\sc i\kern-.025em b}\kern-.08em
    T\kern-.1667em\lower.7ex\hbox{E}\kern-.125emX}}
\begin{document}

\title{Towards Robust Classroom Attendance: A Comprehensive Evaluation of Face Detection and Recognition Models}

\author
{\IEEEauthorblockN{Himani Trivedi}
\IEEEauthorblockA{
\textit{Computer Engineering Department}\\
\textit{LDRP-ITR, Kadi Sarva Vishwavidyalaya}\\
Gandhinagar, Gujarat, India\\
himanitrivedi1994@gmail.com}
\and
\IEEEauthorblockN{Hiren Patel}
\IEEEauthorblockA{
\textit{VSITR, Kadi Sarva Vishwavidyalaya}\\
Kadi, Gujarat, India\\
hbpatel1976@gmail.com}
\and
\IEEEauthorblockN{Ridham Patel\textsuperscript{*}}
\IEEEauthorblockA{
\textit{Information Technology Department}\\
\textit{LDRP-ITR, Kadi Sarva Vishwavidyalaya}\\
Gandhinagar, Gujarat, India\\
ridhampatel2k4@gmail.com}
\and
\IEEEauthorblockN{Krutika Patel}
\IEEEauthorblockA{
\textit{Information Technology Department}\\
\textit{LDRP-ITR, Kadi Sarva Vishwavidyalaya}\\
Gandhinagar, Gujarat, India\\
krutika8114@gmail.com}
\and
\IEEEauthorblockN{Nancy Patel}
\IEEEauthorblockA{
\textit{Information Technology Department}\\
\textit{LDRP-ITR, Kadi Sarva Vishwavidyalaya}\\
Gandhinagar, Gujarat, India\\
nancyp9210@gmail.com}
}
\maketitle

\begin{abstract}
Manual attendance methods, such as paper or register-based systems, take a lot of time, can lead to errors, and are easy to falsify. Face recognition is more reliable, but it frequently struggles in classrooms because lighting and other conditions can vary. Face recognition datasets are designed for regulated environments and do not capture the actual challenges found in classrooms. To address this, a new face detection and recognition dataset, the Visage Face dataset, comprising 16,234 face samples, is proposed for the task of face detection and recognition. The photos are taken from different angles and under varying lighting conditions, with students showing a range of expressions, and some faces partly covered to reflect real-life situations. A YOLO-based system is used to detect faces and tested seven advanced face recognition models with thirteen configurations: LVFace, QCFace, FaceLiVTv2, TopoFR, EdgeFace, TransFace, and GhostFaceNets. Of these, FaceLiVTv2-M performed best, with 99.75\% Top-1/Top-5 accuracy and an inference time of 6.459 ms. These results show that the Visage Face Dataset is a realistic and challenging benchmark for face recognition in classroom attendance.
\end{abstract}

\section{Introduction}

Attendance recording is an essential aspect of academic administration, as regular attendance of students is important in the evaluation of their performance and quality monitoring in educational institutions. Traditional methods, which involve calling out names or using a sign sheet, are still widely used and have been described in the literature as being time-consuming, cumbersome, and prone to the possibility of one student marking the attendance of another who is absent \cite{gomes2020class}.

To overcome these limitations, biometric solutions like RFID, fingerprint, and iris recognition are being used. There are issues with losing, forgetting, misusing, and sharing the RFID cards, but manual data entry can also be prone to mistakes \cite{alniemi2023class}. Fingerprint and iris systems require direct contact with common sensing devices and accurate placement within a limited scanning space, thus posing a risk of transmitting diseases \cite{alniemi2023class}.


The use of facial recognition techniques in automated attendance systems for classrooms has gained considerable attention in recent years, with a significant rise in publications during 2019-2024, especially in the field of deep learning and convolutional neural networks (CNNs) \cite{agustiyar2026face}. However, there are factors such as distance, pose, illumination, occlusion, motion blur, viewpoint, and partial occlusion that can affect the ability to detect and recognise faces in the classroom. Previous studies indicate that the performance of existing models suffers from mask usage due to their poor performance on mostly occluded faces \cite{trivedi2024adapting} \cite{lokhande2023deep}, and real-world datasets have pointed to scale, pose, illumination, and occlusion as major challenges for detection. Therefore, proper attendance is essential to have a good two-stage Face Detection (FD) and Face Recognition (FR) system, and poor face localisation directly impacts face recognition. Although YOLO-based face detection has been studied for low-resolution classroom images, there remains a lack of research evaluating the real-world performance of state-of-the-art face recognition (FR) models under realistic classroom conditions using the same detection framework.

The present work aims to propose a two-stage face detection and recognition system, which motivates the design of a classroom attendance system. The key contributions include:
\begin{itemize}
        \item {Visage Face Dataset (VFD) consisting of 16,234 face samples are proposed for FD and FR. The dataset is designed to mark attendance automatically in educational institutes, under diverse lighting conditions, poses, occlusion, and low resolution. The proposed VFD Dataset comprises two datasets: VFD-Detection and VFD-Recognition.}
        \item This paper compares 12 state-of-the-art detection YOLO models (variants of YOLOv6, YOLOv8, YOLO-NAS, and YOLOv9 variants) to evaluate their performance with the proposed VFD dataset. YOLOv8n is selected for the detection module, delivering a mAP@0.5 of 0.935 and an F1-score of 0.91 with a lightweight architecture.
        \item A thorough assessment of 13 model variants of 7 state-of-the-art face recognition methods (LVFace, TransFace, QCFace, TopoFR, FaceLiVTv2, EdgeFace and GhostFaceNets) is performed on the proposed VFD dataset. FaceLiVTv2-M achieves 99.75\% Top-1 accuracy and 6.459 ms inference time with only 7.02M parameters, showing a good accuracy-inference trade-off.
    \end{itemize}

The subsequent paper is organized as follows: The literature study of the existing face datasets, FD algorithms, and the latest FR models is presented in Section 2. The proposed Visage Face Dataset (VFD) is described in Section 3. The proposed methodology is presented in Section 4, which includes the FD and FR pipeline. Section 5 outlines the experimental results and discussion on the evaluated models, and finally Section 6 outlines the study with future research directions.

\section{Literature Review}
Existing studies related to face-based attendance span three topics: FD and FR datasets, FD models, and FR Architectures. Likewise, this literature review first analyzes FD and FR benchmark datasets, then examines recent FD methods, and thirdly studies the state-of-the-art FR architectures.

\subsection{Literature Review on Benchmark Datasets}
Face databases have been created so far according to the requirements of specific tasks. The Labeled Faces in the Wild (LFW) \cite{huang2008labeled} is a benchmark for unconstrained face verification, but used for verification, and the number of images per identity is limited, making it unsuitable for large-scale evaluation in detection and recognition. However, WIDER FACE \cite{yang2016wider} significantly increased the scale, pose, and occlusion variation of faces to detect, but the annotation was limited to providing support for FD and not FR. On the other hand, VGGFace2 \cite{cao2018vggface2} was created mainly for recognition, with over 3.31 million images of 9,131 identities in various pose and age groups, yet it is not intended to have both detection and recognition tasks.

However, CelebA \cite{liu2015deep} provides over 200,000 face images annotated with face identity, bounding box, landmarks, and attributes, for face detection and recognition. But most of its images are celebrity-oriented web imagery that has significant variation in background and acquisition, as opposed to multi-person classroom imagery. Indian Classroom Face Dataset (ICFD) \cite{bhattacharjee2020d2sc} focuses on recognition in the classroom, where it consists of controlled gallery images and unconstrained classroom probe images of 84 subjects. Despite this, it was used as a low-resolution FR system, with the classroom images serving as probe data rather than as a unified dataset designed to evaluate both FD and FR. In general, most of these datasets focus on specific detection or recognition tasks, and datasets that use the same realistic acquisition setting to simultaneously perform FD and FR are relatively limited.

\subsection{Literature review on FD architectures}
In recent years, FD has been studied in increasingly more complex settings, such as small-scale faces, occlusions, dense scenes, and computation constraints. To overcome the problems of localization failure due to severe background interference, occlusion, and degradation, SFE-DETR \cite{yang2025sfe} developed an improved transformer-based small face detector in complex scenes. FDLite \cite{aggarwal2024fdlite} introduced a compact architecture of 0.26 M parameters and 0.94 GFLOPs for a one-stage face detector that is competitive with the use of computationally heavy backbones. Likewise, Kaparinos et al. \cite{kaparinos2025b} studied structured pruning to reduce detector size while maintaining performance under various levels of difficulty in WIDER FACE. These methods are representative of the general trend of detection techniques that can maintain localization accuracy and yet have lower computational demands for deployment. 

In this regard, YOLO face detectors have attracted the most attention due to their single-stage design, good detection accuracy, and inference efficiency. In order to localize small and partially occluded faces, scale- and occlusion-aware mechanisms were added to the YOLO face detector: YOLO-FaceV2 \cite{yu2024yolo}. These frameworks focus on efficient inference, multi-scale detection, and difficult face instances, which are all highly suitable for real-world applications where numerous faces are required to be localized quickly from a single image.

\subsection{Literature review on FR architectures}
The state-of-the-art FR models have evolved with improved discriminative and structural facial representations. LVFace \cite{you2025lvface} proposed a large vision model based on progressive cluster optimization to boost face representation learning at scale. Meanwhile, TopoFR \cite{dan2024topofr} integrated topology alignment into the face representation learning task using topology-aware hard-sample mining. Image-byte processing, topology-based compression, and structure-guided cross-attention are further areas of study of the FR paradigm explored by TransFace++ \cite{dan2025transface++} for improving accuracy and efficiency.

FR performance also heavily relies on the quality and discriminability of facial samples. QCFace \cite{doan2026qcface} incorporates image-quality control into face representation learning to separate out information of the face from its recognizability and mitigate the negative impact of low-quality images. ExpFace \cite{zheng2025expface} tackles the optimization problem by maximizing class separability and discriminative properties of face embeddings using an exponential angular margin loss.

In addition to high-performance recognition architectures, there are also lightweight models that have gained significance for real-time and resource-constrained applications. To address the computational requirements, GhostFaceNets \cite{alansari2023ghostfacenets} proposed a lightweight model using Ghost operations that achieves competitive recognition performance. Moreover, the Cross Stage Partial bottleneck and channel attention layers, used by GhostFaceNet++ \cite{nachet2025ghostfacenet++}, enhance this efficiency-driven design with the addition of high accuracy and low computation. To achieve efficient FR on edge devices, EdgeFace \cite{george2024edgeface} was designed and offers compact configurations to address different resource limitations.

Recent hybrid architectures also fuse local convolutional feature extraction with global transformer-based representation. For improved recognition accuracy while keeping the computational complexity low, SwiftFaceFormer \cite{luevano2024swiftfaceformer} is designed based on a lightweight CNN-Transformer architecture with knowledge distillation. In the same vein, FaceLiVTv2 \cite{setyawan2026facelivtv2} emphasizes designing a hybrid architecture for mobile face recognition by optimizing the local and global feature processing and lightening the global token interaction. 

These models have been reviewed as a whole and cover a broad range of vision architectures, topology-aware learning, quality-controlled recognition, margin-based optimization, lightweight CNNs, and hybrid CNN-Transformer models. However, they are evaluated mostly according to the conventional FR benchmarks and standardized protocols, which do not necessarily reflect real-world classroom settings. Furthermore, it is difficult to compare the performance of the different architectures directly if they are tested with different datasets and pre-processing pipelines. This encourages a controlled assessment of these state-of-the-art models on a shared classroom dataset and a shared FD and FR pipeline.

\section{Proposed Datasets} \label{sec:proposed dataset}
In the field of face detection and recognition, data quality plays a pivotal role in assessing the model's accuracy and effectiveness. The carefully designed and prepared \textit{VisageFace-Detection (VFD)} dataset provides evidence of the important challenges in this area when targeting faces in classroom environments. The second proposed dataset for the task of FR after FD is \textit{VisageFace-Recognition (VFR)}. Furthermore, this dataset is not only applicable for attendance studies, but it also has a higher utility for FD and FR related tasks.

\subsection{VF-Detection: Proposed Face Detection Dataset}\label{sec:Proposed Face Detection Dataset}

In this study, a face dataset is provided to enhance face detection in an automated attendance system. A total of 801 pictures are captured by hand in classrooms at different schools. These images illustrate real-world scenarios encountered in attendance tracking. The data collection also reflects the increasing use of CCTV footage in schools.

\begin{figure}[!ht]
    \centering
    \includegraphics[width=0.49\textwidth]{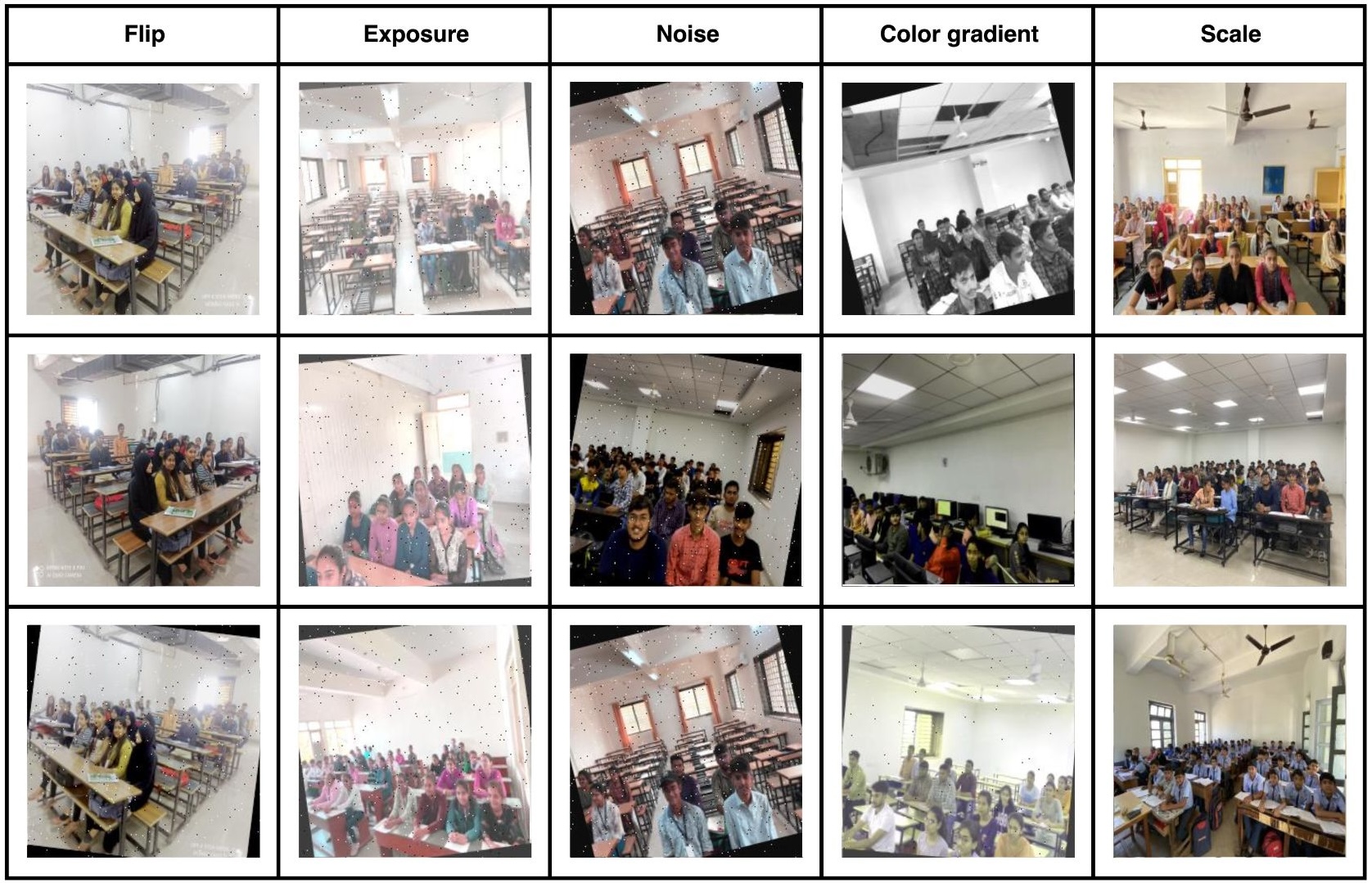}
    \caption{Sample images from the VisageFace-Detection Dataset (original vs. augmented)}
    \label{fig:face_detection_dataset_augmentation}
    \vspace{-15pt}
\end{figure}



\begin{table}[!ht]
    \centering
    \caption{VisageFace Dataset Summary}
    \label{tab:Face_Dataset_summary}
    \renewcommand{\arraystretch}{1.1}
    \footnotesize
    \begin{tabular}{l@{\hspace{7pt}}l@{\hspace{7pt}}l}
        \toprule
        \textbf{Specification} & \textbf{Detection Details} & \textbf{Recognition Details} \\
        \midrule
        Total Images & 3,635 & 3,500 \\
        Dataset Type & Categorical & Categorical \\
        Dataset Size & 224.6 MB & 145.7 MB \\
        Image Dimension & 640 $\times$ 640 pixels & 224 $\times$ 224 pixels \\
        Total Annotations & 96,409 & - \\
        Image Format & JPG & JPEG \\
        Image Sources & Discrete schools, colleges, & Educational institutes \\
        \bottomrule
    \end{tabular}
    
\end{table}
The proposed dataset uses a wide array of augmentation techniques: rotation, scaling, translation, noise augmentation, and flipping, applied separately within the training, validation, and test splits. The images are then manually labeled using the labelImg tool \cite{tzutalin2015labelimg}, and facial instances are carefully labeled. This augmentation process is employed to make the dataset more diverse and more challenging to classify, as shown in Figure \ref{fig:face_detection_dataset_augmentation}, thereby representing changes in lighting, occlusion, pose, and facial expression, different environmental challenges that can be present in the real world. Each image contains approximately 20 student faces, and the dataset can be considered a large dataset for training and testing FD models for the specific needs of student attendance marking in schools. Table ~\ref{tab:Face_Dataset_summary} shows additional information about the proposed dataset.

\subsection{VisageFace-Recognition: Proposed Face Recognition Dataset}

Recently, FR technology has been adopted in various applications, such as attendance management, security, and unlocking personal devices. FR algorithms that are more accurate rely on the quality and diversity of their datasets. To meet this need, and based on the literature discussed earlier, the VisageFace-Recognition (VFR) dataset is introduced with details in Table ~\ref{tab:Face_Dataset_summary}.  It can be employed for training and testing of FR algorithms and to facilitate research on new learning techniques like few-shot learning and deep learning.



As the VFR dataset covers a wide range of facial expressions, lighting conditions, and occlusions, it is a valuable resource for training and testing FR algorithms, thereby enhancing the ability of models to generalize across diverse scenarios. The image is processed using RetinaFace \cite{deng2020retinaface} in order to determine the facial area as well as the landmarks on the face. The landmarks used for alignment are the left eye, right eye, nose tip, left mouth corner, and right mouth corner. These landmarks are then used to align the face using similarity alignment with ArcFace \cite{deng2019arcface}  in order to handle differences in rotations and scales. The original VFR images of 224 × 224 pixels are cropped and resized to 112 × 112 pixels, ensuring a consistent input resolution across all evaluated FR models. This makes sure that FR models have faces that are aligned in terms of orientation and input size but differ in facial expressions and illumination. Furthermore, the availability of 100 different classes allows researchers to explore more detailed aspects of facial feature analysis and classification, improving the effectiveness and reliability of the models. Overall, the VFR dataset is a significant contribution to the field of FR. The dataset has been created in a systematic and organized manner, featuring a large collection of images that can be easily accessed and processed, making it a valuable resource for future research and advancement in this field.

\textbf{Ethics and Privacy:} The face images were collected from students after explaining the purpose of the study. Consent from all participants was obtained prior to data collection for the use of their images in this research. The privacy and confidentiality of the collected data is strictly preserved, and the data is utilised only for this research. Access to the data is limited to the research team.

\section{Methodology}

There are two major phases of this automated attendance system: FD and FR on the proposed VFD dataset. Throughout each step, benchmark models and novel techniques are evaluated for optimal accuracy and speed for real classroom use.

\subsection{Face Detection}

The primary objective of the FD module is to detect students' faces in the classroom and provide precise ROIs for the subsequent FR process. YOLO models, such as YOLOv6, YOLOv8, YOLO-NAS, and YOLOv9, were evaluated on the proposed VFD dataset.

\begin{figure}[!ht]
	\centering
	\includegraphics[width=0.3\textwidth,
        alt={Four sample images from the Visage-Detection dataset showing the model's output. In each classroom scene, multiple student faces are correctly identified with red bounding boxes labeled 'F'.}]{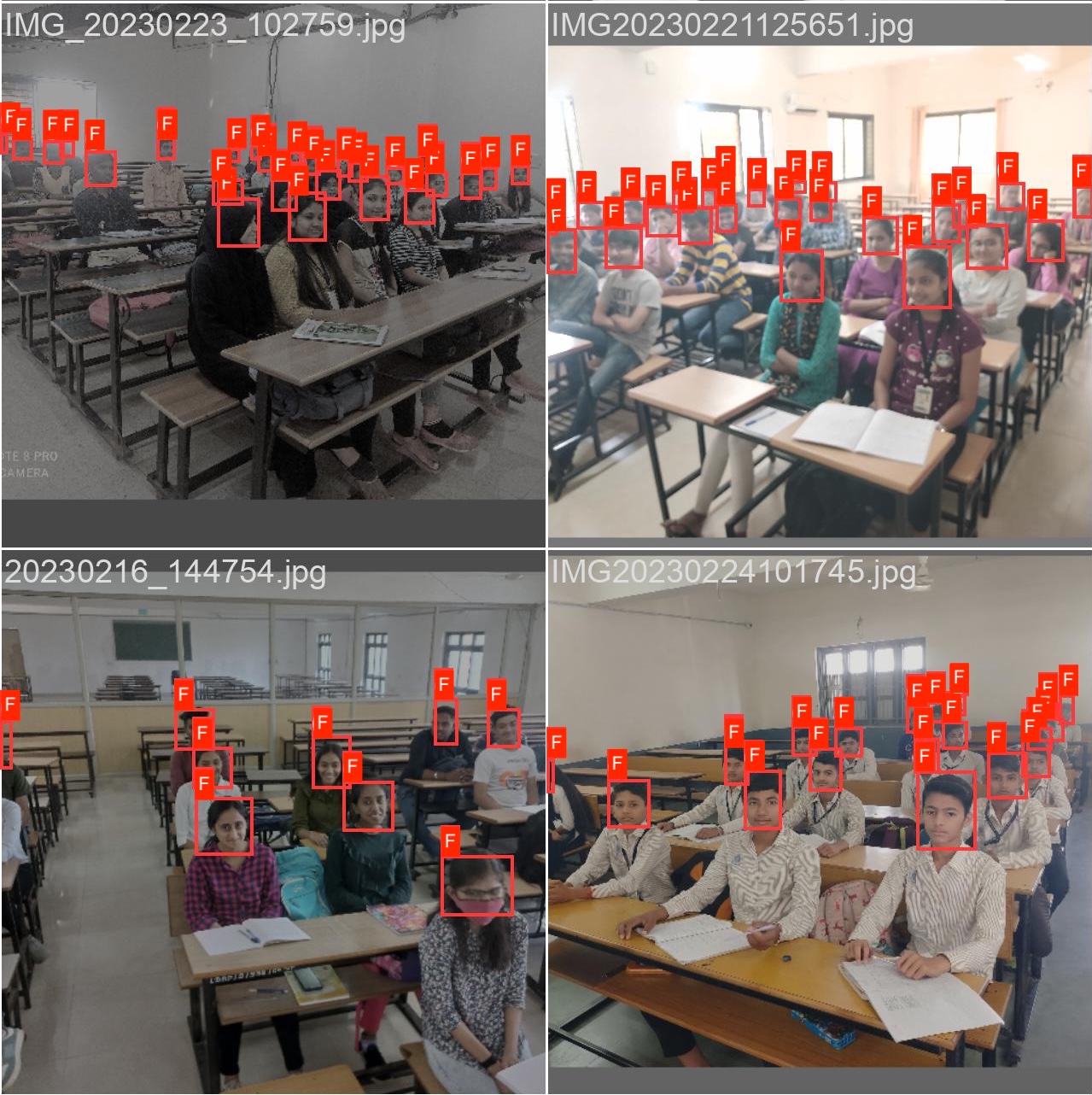}
	\caption{Examples from the proposed VFD dataset under varying classroom conditions, including lighting, occlusion, and student density}
	\label{fig:detection_samples}
    \vspace{-2pt}
\end{figure}

The YOLO models have been trained and evaluated on the VFD dataset, and the results are provided in Figure \ref{fig:detection_samples}. These models are capable of recognising face images under various lighting conditions, at different angles, and of partially covered faces with sunglasses. It is necessary to train the model on a diverse dataset to handle various classroom conditions.

It is found that YOLOv8x achieves the highest mAP; however, due to its high complexity (68.2M parameters), it is inadequate for mobile applications. On the contrary, YOLOv8n has only 3.2M parameters and reaches 93.49\% of mAP. YOLOv8n is selected for the FD step to keep the entire system running without delay.
\begin{figure}[!ht]
	\centering
	\includegraphics[width=1.0\columnwidth,
        alt={A single classroom photograph illustrating a live inference run of the YOLOv8n face detection model. Numerous faces are enclosed in red bounding boxes, each labeled with a confidence score ranging from 0.68 to 0.83.}]{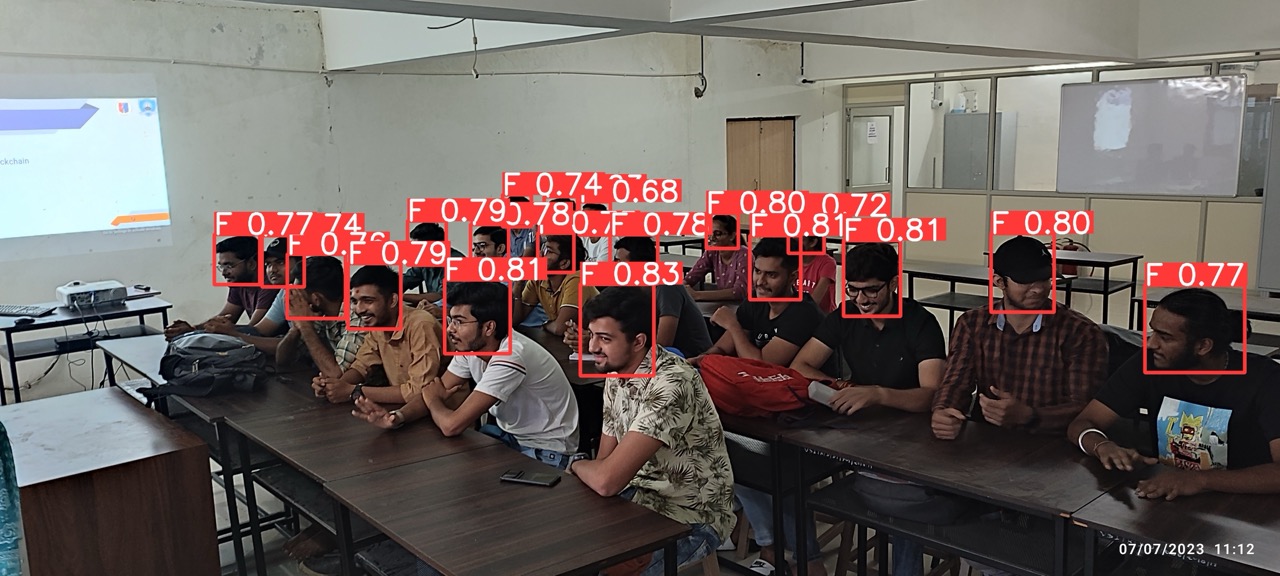}
	\caption{Real-time face detection using the selected YOLOv8n model, with bounding boxes and confidence scores in a classroom setting.}
	\label{fig:live_detection_run}
\end{figure}


The performance of YOLOv8n is illustrated in the real-time detection example shown in Figure \ref{fig:live_detection_run}. As we can see from the example, each face detected by the model is outlined in a bounding box along with its confidence level. The confidence levels are quite high in this example (e.g., 0.77, 0.80, and 0.83), demonstrating the reliability of the algorithm under real-world conditions.

\begin{table*}[t]
\centering
\begin{threeparttable}
\caption{\textit{Evaluation of YOLOv8 Variants on VisageFace-Detection Dataset}}
\
\label{tab:FD_Result_Table_Portrait}
\scriptsize
\setlength{\tabcolsep}{8pt}
\begin{tabular}{lccccccccc}
\toprule
\textbf{Model}
& \textbf{Box Loss} 
& \textbf{Cls Loss} 
& \textbf{Dfl Loss} 
& \textbf{Precision} 
& \textbf{Recall} 
& \textbf{mAP@0.5} 
& \textbf{F1-score}
& \textbf{mAP@.5:.95} 
& \textbf{Params\tnote{1} (M)} \\
\midrule
YOLOv8s & 1.423 & 0.567 & 1.031 & 0.938 & 0.937 & 0.944 & 0.94 & 0.538 & 11.2 \\
YOLOv8m & 1.381 & 0.530 & 1.038 & 0.944 & 0.954 & 0.968 & 0.95 & 0.590 & 25.9 \\
YOLOv8l & 1.258 & 0.498 & 1.004 & 0.964 & \textbf{0.971} & \textbf{0.974} & \textbf{0.97} & 0.604 & 43.7 \\
YOLOv8x & \textbf{1.250} & \textbf{0.495} & \textbf{0.997} & \textbf{0.963} & 0.971 & 0.973 & 0.97 & \textbf{0.616} & 68.2 \\
\rowcolor[HTML]{EFEFEF}
\textbf{YOLOv8n} & 1.418 & 0.784 & 1.059 & 0.932 & 0.886 & 0.935 & 0.91 & 0.530 & \textbf{3.2} \\
\bottomrule
\end{tabular}
\label{tab:FD_Result_Table_Portrait}
\begin{tablenotes}
\footnotesize
\item[1] Number of parameters, in millions.
\end{tablenotes}
\end{threeparttable}
\vspace{-10pt}
\end{table*}

\subsection{Face Recognition}

Face recognition (FR) is implemented in a closed-classroom identification task, where 100 student identities and a total of 3,500 images are split into 2,400 training images, 700 validation images, and 400 test images according to a fixed class-balanced split ratio of 70:20:10. The same training approach is used to fine-tune all evaluated models. Each model generates a 512-dimensional embedding, with a 100-way classification head trained using softmax cross-entropy. AdamW optimiser is utilised with a learning rate of $1\times10^{-4}$, weight decay of $5\times10^{-4}$, a batch size of 64, cosine decay, and a maximum of 100 epochs with early stopping based on validation performance.

For evaluation, all L2-normalised training embeddings are averaged for each identity to create gallery centroids. Each test face is then assigned to the identity with the highest cosine similarity to its corresponding gallery centroid. Recognition performance is measured using Top-1 and Top-5 accuracy, while model efficiency is evaluated using the number of parameters and inference time. To evaluate inference time, the fine-tuned models are exported to ONNX and evaluated using ONNX Runtime with the CUDA execution provider on an NVIDIA Tesla T4. The inference latency is measured with a batch size of one after a warm-up period, and the median latency from 100 runs is used for each model.

\section{Results}

This section provides a detailed analysis of the performance of the face detection and recognition models using different evaluation parameters and determines the optimum model for an automated attendance system in a classroom environment.

\subsection{Face Detection}
The detection module has been tested on several models with special attention to the YOLOv8 architecture and its different variants (n, s, m, l, x). The aim was to find the best compromise between detection accuracy and computation cost in classroom attendance scenarios. 

\begin{figure}[!h]
    \centering
    \includegraphics[width=0.35\textwidth,
        alt={A dual-axis line and scatter plot comparing YOLOv8 model variants. The x-axis shows the models from YOLOv8x to YOLOv8n. The left y-axis shows mAP at 0.5, which slightly decreases as models get smaller. The right y-axis shows the number of parameters in millions, which sharply decreases, illustrating the trade-off between accuracy and model size.}
    ]{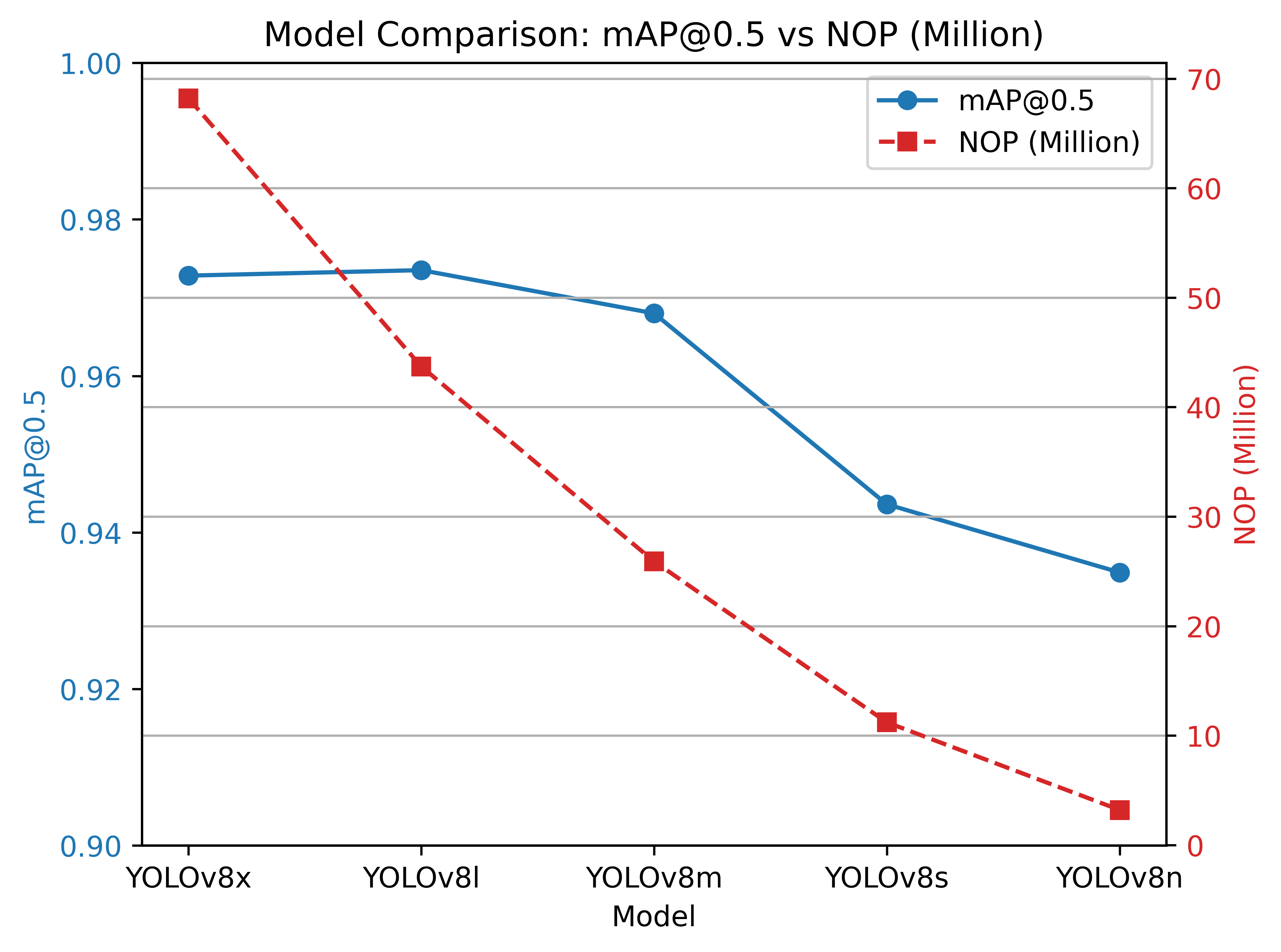}
    \vspace{-5pt}
    \caption{Comparative Graph of Face Detection Models}
    \label{fig:FD_Eval_1}
    \vspace{-5pt}
\end{figure}

\begin{figure}
    \centering
    \includegraphics[width=0.99\linewidth]{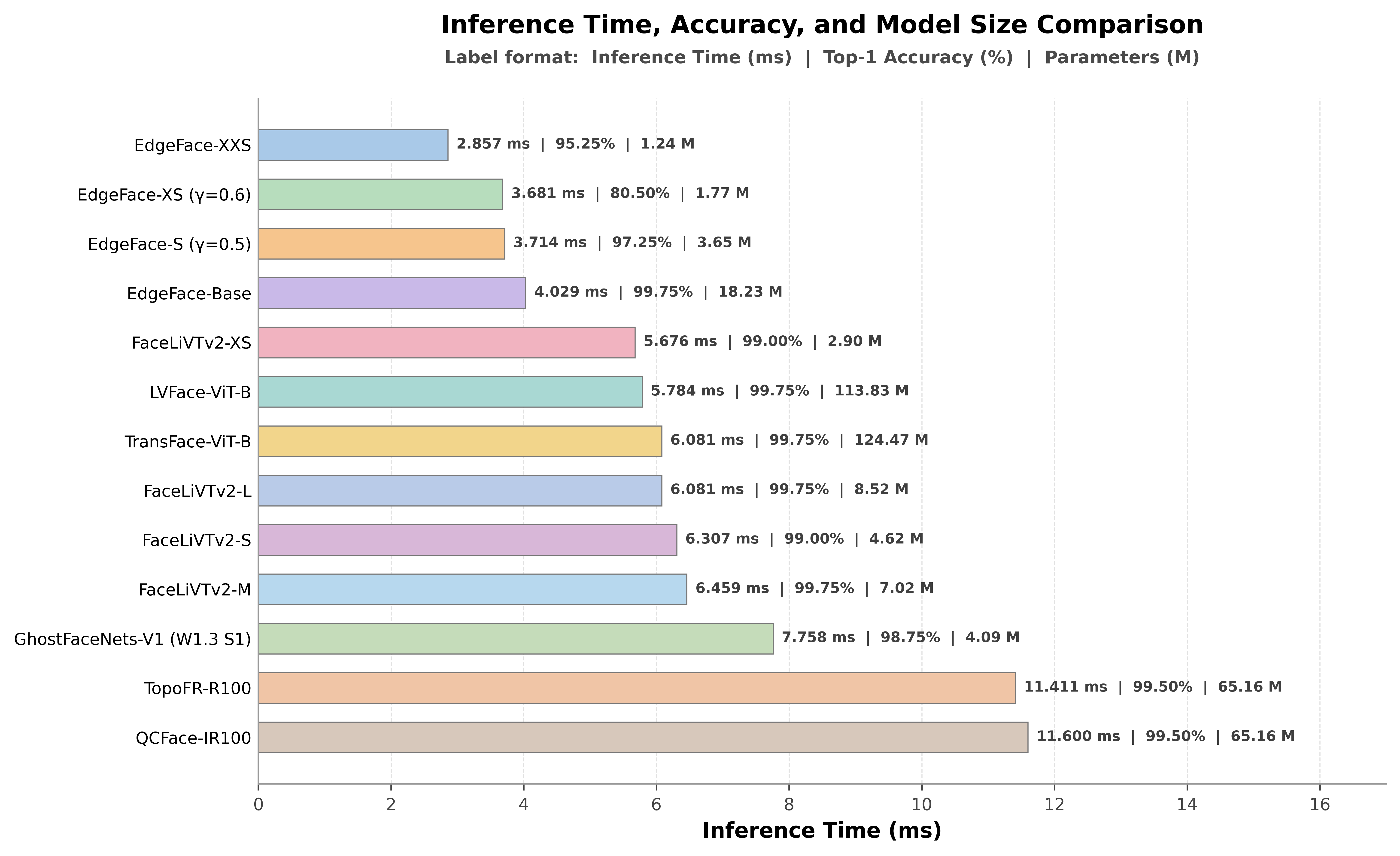}
    \caption{Inference Time, Accuracy, and Model Size Comparison of Evaluated FR Models}
    \label{fig:fr_accuracy_latency}
\end{figure}

YOLOv8n is the most lightweight model in the attendance system, which shows outstanding performance in FD. Despite the small size of its model with only 3.2M parameters, YOLOv8n has a decent mAP@0.5 of 0.935, as seen in Table \ref{tab:FD_Result_Table_Portrait}. This performance is an outstanding efficiency-to-accuracy ratio compared to the other two models mentioned above, YOLOv8l (43.7M) and YOLOv8x (68.2M), having 13.7x and 21.3x more parameters, respectively.

Moreover, YOLOv8n achieves a precision of 0.932, a recall of 0.886, and an F1-score of 0.91. This is a minor compromise when compared to the larger models, but it still has a precision above 93\%, which means that there are fewer false positive detections and the results are reliable. The moderately lower recall means that some faces may not be recalled in some scenarios; however, this is less significant in the case of attendance,e as multiple frames are captured based on a given time period, and the effect of occasional missed recall is reduced. Bigger models yield slightly better results, but the computational benefit of YOLOv8n makes it ideal for such environments that have limited computing resources. 

Inference speed is not specifically mentioned in the comparison table, but is closely related to the number of model parameters. YOLOv8n’s lightweight architecture enables faster processing and supports real-time face detection, with performance depending on the hardware and implementation. This efficiency is important for attendance systems that have to process a video stream from a classroom setting where there are several faces in the videos at the same time. The FD model of the attendance system is selected as the YOLOv8n model. It has outstanding parameter efficiency along with competitive detection performance, resulting in the best trade-off for practical use in educational applications.

\subsection{Face Recognition}

\begin{table}[!h]
\vspace{-12pt}
\centering
\begin{threeparttable}
\caption{\textit{Comparison of FR Models on the VisageFace-Recognition Dataset}}
\label{fr_results}
\scriptsize
\setlength{\tabcolsep}{5.5pt}
\begin{tabular}{llcccc}
\toprule
\textbf{Model}
& \textbf{Version}
& \textbf{Params}
& \textbf{Top-1}
& \textbf{Top-5}
& \textbf{Inference (ms)} \\
\midrule
FaceLiVTv2      & L                 & 8.52   & 99.75 & 100    & 6.081 \\
EdgeFace        & Base              & 18.23  & 99.75 & 99.75  & 4.029 \\
LVFace          & ViT-B             & 113.83 & 99.75 & 100    & 5.784 \\
TopoFR          & R100              & 65.16  & 99.5  & 100    & 11.411 \\
QCFace          & IR100             & 65.16  & 99.5  & 99.75  & 11.6 \\
TransFace       & ViT-B             & 124.47 & 99.75 & 100    & 6.081 \\
FaceLiVTv2      & XS                & 2.9    & 99    & 99.5   & 5.676 \\
FaceLiVTv2      & S                 & 4.62   & 99    & 99.75  & 6.307 \\
GhostFaceNets   & V1 (W1.3 S1)      & 4.09   & 98.75 & 99.75  & 7.758 \\
EdgeFace        & S ($\gamma$=0.5)  & 3.65   & 97.25 & 98.75  & 3.714 \\
EdgeFace        & XXS               & 1.24   & 95.25 & 99.25  & 2.857 \\
EdgeFace        & XS ($\gamma$=0.6) & 1.77   & 80.5  & 95     & 3.681 \\
\rowcolor[HTML]{EFEFEF}
\textbf{FaceLiVTv2} & \textbf{M} & \textbf{7.02} & \textbf{99.75} & \textbf{99.75} & \textbf{6.459} \\
\bottomrule
\end{tabular}
\label{tab:fr_results}
\begin{tablenotes}
\footnotesize
\item W = Width, S = Stride, $\gamma$ = LoRaLin rank ratio.
\end{tablenotes}
\end{threeparttable}
\end{table}

The results of the FR are shown in Tab. \ref{tab:fr_results}. The highest Top-1 accuracy of 99.75\% is obtained by five configurations: FaceLiVTv2-M, FaceLiVTv2-L, EdgeFace-Base, LVFace, and TransFace. FaceLiVTv2-XS and S need just 2.99M and 4.65M parameters and attain 99.00\% accuracy. EdgeFace variants have more scattered results,s with EdgeFace-S, XXS, and XS having 97.25\%, 95.25\%, and 80.50\% Top-1 accuracy, respectively. Overall, the results show high recognition performance for most of the tested architectures on the VFD test set.

The accuracy-latency relationship is shown in Fig.~\ref{fig:fr_accuracy_latency}. Although EdgeFace-XXS has a lower Top-1 accuracy of 95.25\%, it achieves the lowest overall latency of 2.857~ms. The models achieving the highest Top-1 accuracy of 99.75\% have inference latencies ranging from 4.029~ms to 6.459~ms. However, TopoFR and QCFace take 11.411~ms and 11.600~ms, respectively, to reach an accuracy of 99.50\% respectively. The results in this paper show that the efficiency of the model varies significantly between the evaluated models and that the accuracy of the recognition and the response time in the classroom should also be considered when choosing a model.

\section*{Conclusion}
Two face datasets, VisageFace-Detection and VisageFace-Recognition, have been created to tackle challenges in actual classroom attendance scenarios. The VisageFace Dataset includes 16,234 labeled face samples from classroom settings, showing a range of poses, lighting, occlusions, positions, and expressions. Tests with top recognition models show that FaceLiVTv2-M reaches 99.75\% Top-1 accuracy, uses 7.02M parameters, and processes each image in 6.46~ms. This depicts a good balance between accuracy and efficiency. The results also show that smaller or combined models can perform as well as much larger ones, making them suitable for real-time classroom attendance systems. Future research directions comprise the development of a classroom-ready recognition system, the expansion of both datasets to incorporate data from additional educational institutions and more challenging environmental conditions, and the integration of detection and recognition modules into a unified, real-time attendance system optimised for efficient operation on low-resource devices.
\section*{Acknowledgment}
This work was supported by the M. M. Patel Research Cell, Kadi Sarva Vishwavidyalaya under Grant MMPRC/2022-23/13.

\vspace{-7.5pt}
\bibliographystyle{ieeetr}
\bibliography{bibliography}

@inproceedings{gomes2020class,
  title={Class attendance management system using facial recognition},
  author={Gomes, Clyde and Chanchal, Sagar and Desai, Tanmay and Jadhav, Dipti},
  booktitle={ITM web of conferences},
  volume={32},
  pages={02001},
  year={2020},
  organization={EDP Sciences}
}

@article{alniemi2023class,
  title={Class attendance system based on face recognition},
  author={Alniemi, Omar and Mahmood, Hanaa F},
  journal={Revue d’Intelligence Artificielle},
  volume={37},
  number={5},
  pages={1245--1253},
  year={2023}
}

@article{agustiyar2026face,
  title={Face Recognition for Attendance Systems: A Bibliometric Review of Research Trends and Opportunities},
  author={Agustiyar, Agustiyar and Isnanto, R Rizal and Widodo, Catur Edi},
  journal={Jurnal Sisfokom (Sistem Informasi dan Komputer)},
  volume={15},
  number={1},
  pages={8--13},
  year={2026}
}

@article{bhattacharjee2020d2sc,
  title={D2SC-GAN: dual deep-shallow channeled generative adversarial network, for resolving low-resolution faces for recognition in classroom scenarios},
  author={Bhattacharjee, Avishek and Das, Sukhendu},
  journal={IEEE Transactions on Biometrics, Behavior, and Identity Science},
  volume={2},
  number={3},
  pages={223--234},
  year={2020},
  publisher={IEEE}
}

@article{trivedi2024adapting,
  title={Adapting face recognition to the masked world: leveraging deep attention networks with a custom dataset},
  author={Trivedi, Himani and Goyani, Mahesh},
  journal={International Journal of Computers and Applications},
  volume={46},
  number={6},
  pages={427--441},
  year={2024},
  publisher={Taylor \& Francis}
}

@inproceedings{huang2008labeled,
  title={Labeled faces in the wild: A database forstudying face recognition in unconstrained environments},
  author={Huang, Gary B and Mattar, Marwan and Berg, Tamara and Learned-Miller, Eric},
  booktitle={Workshop on faces in'Real-Life'Images: detection, alignment, and recognition},
  year={2008}
}

@inproceedings{yang2016wider,
  title={Wider face: A face detection benchmark},
  author={Yang, Shuo and Luo, Ping and Loy, Chen-Change and Tang, Xiaoou},
  booktitle={Proceedings of the IEEE conference on computer vision and pattern recognition},
  pages={5525--5533},
  year={2016}
}

@inproceedings{cao2018vggface2,
  title={Vggface2: A dataset for recognising faces across pose and age},
  author={Cao, Qiong and Shen, Li and Xie, Weidi and Parkhi, Omkar M and Zisserman, Andrew},
  booktitle={2018 13th IEEE international conference on automatic face \& gesture recognition (FG 2018)},
  pages={67--74},
  year={2018},
  organization={IEEE}
}

@inproceedings{liu2015deep,
  title={Deep learning face attributes in the wild},
  author={Liu, Ziwei and Luo, Ping and Wang, Xiaogang and Tang, Xiaoou},
  booktitle={Proceedings of the IEEE international conference on computer vision},
  pages={3730--3738},
  year={2015}
}

@article{yang2025sfe,
  title={SFE-DETR: An Enhanced Transformer-Based Face Detector for Small Target Faces in Open Complex Scenes},
  author={Yang, Chenhao and Jiang, Yueming and Song, Chunyan},
  journal={Sensors},
  volume={26},
  number={1},
  pages={125},
  year={2025},
  publisher={MDPI}
}

@article{aggarwal2024fdlite,
  title={FDLite: A single stage lightweight face detector network},
  author={Aggarwal, Yogesh and Guha, Prithwijit},
  journal={arXiv preprint arXiv:2406.19107},
  year={2024}
}

@inproceedings{kaparinos2025b,
  title={B-fpgm: Lightweight face detection via bayesian-optimized soft fpgm pruning},
  author={Kaparinos, Nikolaos and Mezaris, Vasileios},
  booktitle={2025 IEEE/CVF Winter Conference on Applications of Computer Vision Workshops (WACVW)},
  pages={844--853},
  year={2025},
  organization={IEEE}
}

@article{yu2024yolo,
  title={Yolo-facev2: A scale and occlusion aware face detector},
  author={Yu, Ziping and Huang, Hongbo and Chen, Weijun and Su, Yongxin and Liu, Yahui and Wang, Xiuying},
  journal={Pattern Recognition},
  volume={155},
  pages={110714},
  year={2024},
  publisher={Elsevier}
}

@inproceedings{you2025lvface,
  title={LVFace: Progressive cluster optimization for large vision models in face recognition},
  author={You, Jinghan and Li, Shanglin and Sun, Yuanrui and Wei, Jiangchuan and Guo, Mingyu and Feng, Chao and Ran, Jiao},
  booktitle={2025 IEEE/CVF International Conference on Computer Vision (ICCV)},
  pages={11840--11849},
  year={2025},
  organization={IEEE}
}

@article{dan2024topofr,
  title={Topofr: A closer look at topology alignment on face recognition},
  author={Dan, Jun and Liu, Yang and Deng, Jiankang and Xie, Haoyu and Li, Siyuan and Sun, Baigui and Luo, Shan},
  journal={Advances in Neural Information Processing Systems},
  volume={37},
  pages={37213--37240},
  year={2024}
}

@article{dan2025transface++,
  title={TransFace++: Rethinking the Face Recognition Paradigm with a Focus on Accuracy, Efficiency, and Security},
  author={Dan, Jun and Liu, Yang and Sun, Baigui and Deng, Jiankang and Luo, Shan},
  journal={IEEE Transactions on Pattern Analysis and Machine Intelligence},
  year={2025},
  publisher={IEEE}
}

@inproceedings{doan2026qcface,
  title={QCFace: Image Quality Control for boosting Face Representation \& Recognition},
  author={Doan-Ngo, Duc-Phuong and Diep, Thanh-Dang and Nguyen-Duc, Thanh and Le, Thanh-Sach and Thoai, Nam},
  booktitle={2026 IEEE/CVF Winter Conference on Applications of Computer Vision (WACV)},
  pages={1501--1511},
  year={2026},
  organization={IEEE}
}

@article{zheng2025expface,
  title={ExpFace: Exponential Angular Margin Loss for Deep Face Recognition},
  author={Zheng, Jinhui and Gong, Xueyuan},
  journal={arXiv preprint arXiv:2509.19753},
  year={2025}
}

@article{alansari2023ghostfacenets,
  title={Ghostfacenets: Lightweight face recognition model from cheap operations},
  author={Alansari, Mohamad and Hay, Oussama Abdul and Javed, Sajid and Shoufan, Abdulhadi and Zweiri, Yahya and Werghi, Naoufel},
  journal={IEEE Access},
  volume={11},
  pages={35429--35446},
  year={2023},
  publisher={IEEE}
}

@article{nachet2025ghostfacenet++,
  title={GhostFaceNet++: boosting efficiency and accuracy via CSP bottlenecks and Channel Attention},
  author={Nachet, Randa and Garrig{\'o}s, Javier and Stambouli, Tarik Boudghene},
  journal={Journal of Real-Time Image Processing},
  volume={22},
  number={6},
  pages={201},
  year={2025},
  publisher={Springer}
}

@article{george2024edgeface,
  title={Edgeface: Efficient face recognition model for edge devices},
  author={George, Anjith and Ecabert, Christophe and Shahreza, Hatef Otroshi and Kotwal, Ketan and Marcel, S{\'e}bastien},
  journal={IEEE Transactions on Biometrics, Behavior, and Identity Science},
  volume={6},
  number={2},
  pages={158--168},
  year={2024},
  publisher={IEEE}
}

@inproceedings{luevano2024swiftfaceformer,
  title={Swiftfaceformer: An efficient and lightweight hybrid architecture for accurate face recognition applications},
  author={Luevano, Luis S and Mart{\'\i}nez-D{\'\i}az, Yoanna and M{\'e}ndez-V{\'a}zquez, Heydi and Gonz{\'a}lez-Mendoza, Miguel and Frey, Davide},
  booktitle={International Conference on Pattern Recognition},
  pages={244--258},
  year={2024},
  organization={Springer}
}

@article{setyawan2026facelivtv2,
  title={FaceLiVTv2: An Improved Hybrid Architecture for Efficient Mobile Face Recognition},
  author={Setyawan, Novendra and Sun, Chi-Chia and Hsu, Mao-Hsiu and Kuo, Wen-Kai and Hsieh, Jun-Wei},
  journal={IEEE Transactions on Biometrics, Behavior, and Identity Science},
  year={2026},
  publisher={IEEE}
}

@inproceedings{lokhande2023deep,
  title={Deep learning model based face mask detection for automated mandation},
  author={Lokhande, Pratham and Surati, Shivangi and Trivedi, Himani and Shrimali, Bela},
  booktitle={2023 International Conference on Advancement in Computation \& Computer Technologies (InCACCT)},
  pages={166--170},
  year={2023},
  organization={IEEE}
}

@misc{tzutalin2015labelimg,
	author = {Tzutalin},
	title = {LabelImg},
	howpublished = {Free Software: MIT License},
	year = {2015},
	url = {https://github.com/tzutalin/labelImg}
}

@inproceedings{deng2019arcface,
  title     = {ArcFace: Additive Angular Margin Loss for Deep Face Recognition},
  author    = {Deng, Jiankang and Guo, Jia and Xue, Niannan and Zafeiriou, Stefanos},
  booktitle = {Proceedings of the IEEE/CVF Conference on Computer Vision and Pattern Recognition (CVPR)},
  pages     = {4690--4699},
  year      = {2019}
}

@inproceedings{deng2020retinaface,
  title     = {RetinaFace: Single-Shot Multi-Level Face Localisation in the Wild},
  author    = {Deng, Jiankang and Guo, Jia and Ververas, Evangelos and Kotsia, Irene and Zafeiriou, Stefanos},
  booktitle = {Proceedings of the IEEE/CVF Conference on Computer Vision and Pattern Recognition (CVPR)},
  pages     = {5203--5212},
  year      = {2020}
}

\end{document}